# GPU Memory Requirement Prediction for Deep Learning Task Based on Bidirectional Gated Recurrent Unit Optimization Transformer

Chao Wang*
*Department of Computer Science*
Rice University
Houston, United States
zylj2020@outlook.com

Zhizhao Wen
*Department of Computer Science*
Rice University
Houston, United States
zhizhaowen@alumni.rice.edu

Ruoxin Zhang
*Department of Computer Science*
Rice University
Houston, United States
lz37@alumni.rice.edu

Puyang Xu
*Department of Electrical and Computer Engineering*
Duke University
Durham, United States
rivocxu@gmail.com

Yifan Jiang
*Department of Electrical and Computer Engineering*
Duke University
Durham, United States
Yifan.jiang999@gmail.com

*Abstract*—In response to the increasingly critical demand for accurate prediction of GPU memory resources in deep learning tasks, this paper deeply analyzes the current research status and innovatively proposes a deep learning model that integrates bidirectional gated recurrent units (BiGRU) to optimize the Transformer architecture, aiming to improve the accuracy of memory demand prediction. To verify the effectiveness of the model, a carefully designed comparative experiment was conducted, selecting four representative basic machine learning models: decision tree, random forest, Adaboost, and XGBoost as benchmarks. The detailed experimental results show that the BiGRU Transformer optimization model proposed in this paper exhibits significant advantages in key evaluation indicators: in terms of mean square error (MSE) and root mean square error (RMSE), the model achieves the lowest value among all comparison models, and its predicted results have the smallest deviation from the actual values; In terms of mean absolute error (MAE) and coefficient of determination ($R^2$) indicators, the model also performs well and the results are balanced and stable, with comprehensive predictive performance far exceeding the benchmark machine learning methods compared. In summary, the Transformer model based on bidirectional gated recurrent unit optimization successfully constructed in this study can efficiently and accurately complete GPU memory demand prediction tasks in deep learning tasks, and its prediction accuracy has been significantly improved compared to traditional machine learning methods. This research provides strong technical support and reliable theoretical basis for optimizing resource scheduling and management of deep learning tasks, and improving the utilization efficiency of computing clusters.

*Keywords—GPU memory resources, Bidirectional gate controlled loop unit, Transformer*

## I. Introduction

The accurate prediction of GPU memory requirements in deep learning tasks is a key challenge in the current field of artificial intelligence infrastructure management and optimization, and its research background is deeply rooted in the explosive growth of deep learning model scale and complexity [1]. With the number of parameters such as Transformer architecture, Large Language Models (LLMs), and diffusion models reaching billions or even trillions, the memory consumption during model training and inference has become a core bottleneck that restricts computing efficiency. Insufficient video memory not only leads to training interruptions, limited batch processing sizes, and low utilization of computing resources, but also significantly increases the high communication and hardware acquisition costs in distributed training. The traditional "trial and error method" or rough estimation based on experience is inadequate when facing modern models with complex and variable structures and dynamically generated computational graphs. There is an urgent need for more intelligent and automated memory demand prediction methods to guide efficient resource allocation, task scheduling, model design, and distributed strategy formulation.

Machine learning algorithms play a core driving role in GPU memory demand prediction for deep learning tasks, and their core value lies in learning and modeling the complex nonlinear mapping relationship between memory consumption and numerous influencing factors from historical data [2]. Recent advances in neural network architectures have demonstrated significant potential in modeling complex nonlinear relationships in engineering applications, particularly in dynamic analysis tasks [3] and robot systems [4, 5]. The key to predictive models lies in carefully designed feature engineering: input features typically include model architecture parameters, operation types, optimizer and its state, batch size, sequence length, accuracy of activated data types, and whether to enable memory optimization techniques. By utilizing these features, machine learning algorithms or recurrent neural networks that are better at capturing temporal dependencies can construct high-precision regression models to predict peak video memory usage under specific configurations [6]. Graph neural networks (GNNs) are particularly suitable for encoding the computational graph structure itself as input to model the joint effects of nodes and edges on video memory in the graph. These learned models can replace or significantly enhance traditional static estimation methods based on theoretical formulas (such as accumulating based on parameter quantities, activation quantities, and optimizer states), providing more realistic dynamic prediction results, especially in effectively dealing with the uncertainty caused by dynamic changes in computational graphs.

Although machine learning methods have significantly improved the accuracy of predictions, this field still faces many challenges and is a key direction for future research. The core challenge lies in the extreme dynamism of model behavior: dynamic computation graphs, sparse activation patterns, underlying implementation details of specific frameworks/operator libraries, and subtle differences in hardware drivers all make it exceptionally difficult to build fully universal prediction models. Current research is focused on improving the generalization ability of models, enabling them to be transferred to unseen model architectures or tasks, which may require the integration of meta learning or transfer learning techniques [7]. Meanwhile, exploring how to seamlessly integrate prediction results into automated machine learning processes, achieving neural architecture automation (NAS) under memory constraints, or for real-time guidance of distributed training frameworks for optimal parallel partitioning and resource scheduling, is a highly valuable frontier direction. This article proposes a deep learning algorithm based on bidirectional gated recurrent unit optimization Transformer for GPU memory demand prediction in deep learning tasks, in response to the current research status.

## II. DATA FROM DATA ANALYSIS

The dataset selected in this article is a private dataset, which is a deep learning task resource requirement dataset that can be used to optimize GPU resource sharing and cluster scheduling. It contains 452 samples, each record representing a feature of a deep learning training or inference task and its corresponding GPU resource requirement. The core features of the dataset include input variables such as task type, model architecture, input dimension, batch size, network layers, parameter count, and computational accuracy, while accurately recording the GPU memory required for each task as the prediction target. Select some of the datasets for display, as shown in Table 1.

TABLE I. SOME OF THE DATA

| Input dim | Memory usage mb | Model arch | Num layers | Num parameters | Precision encoded | Parameters per layer |
|---|---|---|---|---|---|---|
| 128 | 27416 | VGG16 | 83 | 190063585 | 2 | 2.2363 |
| 442 | 48000 | BERT | 75 | 8817833 | 0 | 0.1148 |
| 320 | 48000 | VGG16 | 88 | 28450395 | 0 | 0.3157 |
| 285 | 48000 | YOLO v4 | 54 | 77751774 | 1 | 1.4061 |
| 117 | 6989 | YOLO v4 | 89 | 13406168 | 0 | 0.1471 |
| 353 | 48000 | U-Net | 85 | 178468 | 2 | 0.0021 |
| 41 | 15751 | BERT | 15 | 1706766 | 1 | 0.1111 |
| 36 | 32422 | GAN | 65 | 1396224 | 1 | 0.0210 |

## III. METHOD

### A. Transformer

Transformer is a revolutionary deep learning architecture that has fundamentally changed fields such as natural language processing. The network structure of Transformer is shown in Figure 1, and its core lies in completely abandoning traditional recurrent neural networks (RNNs) and convolutional neural networks (CNNs), and instead relying on self attention mechanisms to model the dependency relationships between elements in sequence data, regardless of their distance. Transformer adopts a standard encoder decoder structure. The encoder is composed of multiple identical layers stacked together, each layer containing a multi head self attention sublayer and a feedforward neural network sublayer, with residual connections layer normalization applied around each sublayer [8]. The self attention mechanism calculates the "relevance" score of each element in the sequence relative to all other elements in the sequence, and uses these scores to weight the aggregated value vector, thereby generating a new representation for each element that integrates the global context. The multi head mechanism allows the model to learn information in parallel from different representation subspaces [9]. The decoder structure is similar to an encoder, but an additional encoder decoder attention layer is added between its multi head self attention sublayer and the encoder output. The mask ensures that the decoder can only focus on the previously generated output positions in the sequence when generating the output at the current position, maintaining the autoregressive properties. The encoder decoder attention layer allows the decoder to focus on the complete input sequence representation output by the encoder. The positional information is injected into the input embedding through explicit positional encoding, which compensates for the model's lack of perceptual order ability.

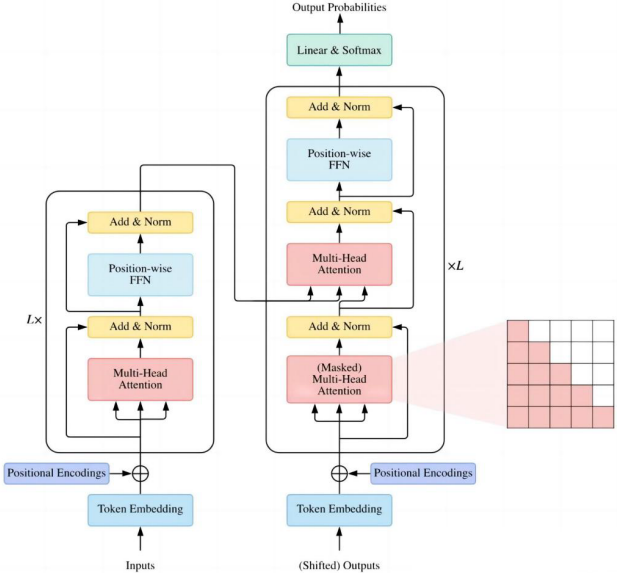

Fig. 1. The network structure of Transformer.

## B. BiGRU

BiGRU (Bidirectional Gated Recurrent Unit) is an extended architecture of GRU (Gated Recurrent Unit) specifically designed to more effectively capture contextual dependencies in sequence data. The network structure of BiGRU is shown in Figure 2, and its core principle is based on the standard GRU. Recent research has shown that optimizing bi-directional gated loop cells with multi-head attention mechanisms can significantly enhance performance in classification tasks, demonstrating the potential for similar improvements in regression applications [10]. GRU solves the gradient vanishing/exploding problem of traditional RNNs by introducing two sophisticated gating mechanisms - update gate and reset gate - and better controls the flow and memory of information [11].

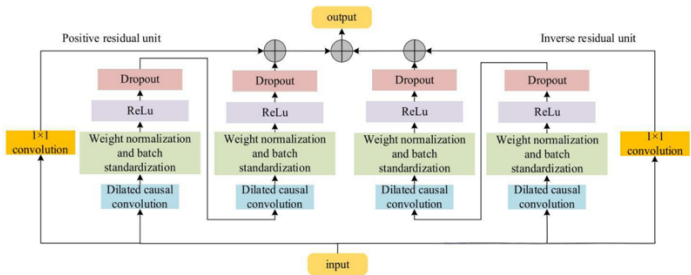

Fig. 2. The network structure of BiGRU

The update gate determines how much of the current input information and the previous hidden state information are retained and fused to form new candidate hidden states; The reset gate controls how much information in the previous hidden state needs to be "forgotten" or reset to calculate new candidate states. Ultimately, the new hidden state is the weighted sum of the previous hidden state and the current candidate state, with the weight controlled by the update gate. This gating structure enables GRU to adaptively learn long-term and short-term dependencies, and has stronger modeling capabilities than the base RNN.

## C. Transformer based on bidirectional gating loop unit optimization

Although Transformer achieves global context modeling through self attention mechanism, it has significant shortcomings in sequence feature processing for GPU memory demand prediction. From the perspective of local dependency capture, self attention performs global correlation calculations on all elements, resulting in low processing efficiency for strong local correlation patterns such as "adjacent network layers-parameter count per layer-local memory consumption" - even if the correlation between adjacent features is much higher than that of distant features, it still requires the same amount of computing resources, leading to redundancy and loss of accuracy in local information modeling. In terms of positional information processing, Transformer relies on sine cosine static encoding or shallow learning positional encoding, which cannot dynamically adapt to complex scenes predicted by GPU memory: the fixed periodic pattern of the former is difficult to reflect the nonlinear positional impact of increasing batch size on memory consumption, while the latter can only learn surface positional correlations and cannot capture the positional feature weight shift caused by model architecture differences, thereby affecting the effective transmission of key positional information in the sequence.

BiGRU precisely compensates for the aforementioned shortcomings of Transformers through its bidirectional gating structure and dynamic temporal modeling capability. Its bidirectional design is divided into a forward GRU layer and a backward GRU layer: the forward layer gradually transmits information from the beginning to the end of the sequence, focusing on capturing the cumulative impact of historical features; The backward layer traverses backwards from the end of the sequence, focusing on the constraints of future features on the current node. After bidirectional information fusion, it can generate a more complete local context representation. The core gating mechanism further optimizes information filtering: the update gate dynamically determines the fusion ratio of "current input features" and "historical hidden states" through sigmoid activation; The reset gate suppresses irrelevant historical information to avoid noise interference in local dependency modeling. In addition, the cyclic structure of BiGRU naturally contains temporal order, which allows for dynamic learning of the intrinsic correlation between "feature input order" and memory consumption without the need for additional positional encoding. After embedding BiGRU into the Transformer, BiGRU first preprocesses the sequence to generate optimized local feature representations, and then inputs them into the Transformer's self attention layer, allowing the Transformer to focus on global correlation modeling. This not only reduces the computational burden of long sequences, but also enhances the accuracy of capturing local key information, providing a dual guarantee for accurate prediction of GPU memory requirements.

## IV. RESULT

In the experiment, we used a Transformer model based on bidirectional gated recurrent unit optimization for regression tasks. The specific configuration includes 6 layers of Transformer encoder, 512 hidden unit sizes per layer, 8 attention heads, and embedding a bidirectional GRU layer for optimization. The hidden unit size is 256, and the dropout rate is set to 0.1 to alleviate overfitting. Optimization is based on Adam optimizer; The hardware environment is NVIDIA GeForce RTX 3090 GPU (24GB video memory) and Intel Core i9-10900K processor, the software platform is Matlab R2024a, and the system memory is 32GB, ensuring efficient operation of the experiment.

In comparative experiments, this article uses four models: decision tree, random forest, Adaboost, and XGBoost. In terms of evaluation parameters, this article selects MSE, RMSE, MAE, MAPR, and R2.

Output the comparison of various parameters between each comparative experimental model and Our model, as shown in Table 2.

TABLE II. THE COMPARISON OF VARIOUS PARAMETERS BETWEEN EACH COMPARATIVE EXPERIMENTAL MODEL AND OUR MODEL

| Model | MSE | RMSE | MAE | MAPE | $R^2$ |
|---|---|---|---|---|---|
| Decision tree | 1103.925 | 33.225 | 4.195 | 70.194 | -6.003 |
| Random forest | 215.868 | 14.692 | 3.255 | 48.53 | -0.402 |
| Adaboost | 153.575 | 12.393 | 1.871 | 63.858 | 0.445 |
| XGBoost | 297.774 | 17.256 | 2.299 | 25.269 | 0.729 |
| Our model | 81.771 | 9.043 | 3.616 | 261.029 | 0.408 |

According to Table 2, our model has the lowest MSE, followed by the Adaboost model; On RMSE, Our model is also the lowest; Our model performs relatively evenly on MAE and $R^2$. In summary, the Transformer model based on bidirectional gated recurrent unit optimization proposed in this article can effectively predict GPU memory requirements in deep learning tasks, and has much higher prediction accuracy compared to basic machine learning algorithms.

Select the comparative test results of MSE and RMSE for bar chart display, as shown in Figures 3 and 4.

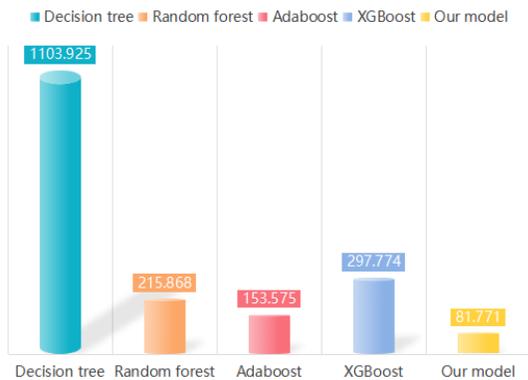

Fig. 3. The comparative test results of MSE.

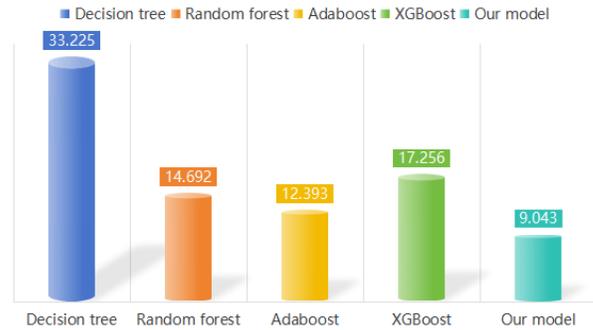

Fig. 4. The comparative test results of MSE.

TABLE III. COMPARATIVE RESULTS OF ABLATION EXPERIMENTS

| Model | MSE | RMSE | MAE | MAPE | $R^2$ |
|---|---|---|---|---|---|
| Transformer | 95.328 | 9.764 | 3.952 | 255.783 | 0.342 |
| Our model | 81.771 | 9.043 | 3.616 | 261.029 | 0.408 |

From the numerical comparison, our model has better overall performance than Transformer: in terms of core error indicators, Transformer's MSE (95.328) is 13.557 higher than our model (81.771), RMSE (9.764) is 0.721 higher than our model (9.043), and MAE (3.952) is 0.336 higher than our model (3.616). All three error indicators are significantly larger, indicating that the deviation between Transformer's predicted values and true values is more significant; In terms of model fitting ability, the $R^2$ of Transformer (0.342) is 0.066 lower than that of Our model (0.408), indicating its weak ability to explain data patterns; Only in terms of MAPE (Mean Absolute Percentage Error), Transformer (255.783) is slightly lower than Our model (261.029), with a difference of 4.246, but both are at a very high MAPE level, proving the effectiveness of our model.

V. CONCLUSION

In response to the current research status of GPU memory demand prediction in deep learning tasks, this study proposes an innovative deep learning algorithm. This algorithm optimizes the Transformer model by introducing a Bidirectional Gated Recurrent Unit (BiGRU), aiming to more accurately predict the memory consumption during deep learning task execution. In order to comprehensively evaluate the performance of our model, we conducted systematic comparative experiments with four classic machine learning models: decision tree, random forest, Adaboost, and XGBoost. The experimental results show that the optimization model proposed in this paper exhibits significant advantages in key performance evaluation indicators. Specifically, in terms of measuring the mean square error (MSE) and root mean square error (RMSE) indicators of prediction error, Our model achieved the lowest value and the highest prediction accuracy, with Adaboost model performing second best in MSE; Meanwhile, in terms of mean absolute error (MAE) and coefficient of determination ($R^2$) indicators, our model's performance is also quite balanced and excellent, demonstrating the robustness of the overall predictive ability of the model.

Based on the comprehensive experimental results, the Transformer model proposed in this study, which is based on bidirectional gated recurrent unit optimization, performs well in predicting GPU memory requirements for deep learning tasks. Its prediction accuracy significantly exceeds that of various machine learning algorithms used as a comparative basis. This model can effectively capture the complex temporal dependencies of video memory usage patterns, providing a more reliable and accurate tool for estimating video memory demand for resource planning, scheduling optimization, and cost control in deep learning workloads. Therefore, the results of this study not only improve the accuracy of GPU memory prediction, but also provide key technical support for efficient utilization of expensive computing hardware resources and optimization of deep learning system performance.

## VI. DISCUSS

This study innovatively integrates BiGRU into the Transformer architecture, effectively addressing the shortcomings of Transformer in capturing local dependencies and dynamic location information, and making valuable contributions to solving the key challenge of GPU memory demand prediction in deep learning tasks. Specifically, BiGRU's bidirectional gating mechanism enhances the modeling of local correlations and inherently encodes temporal order without relying on static positional encoding, while Transformer's self attention mechanism maintains global context modeling. The experimental results further validated the superiority of the model: among the compared traditional machine learning models (decision tree, random forest, Adaboost, XGBoost), it achieved the lowest MSE (81.771) and RMSE (9.043), outperforming Transformer in core error metrics (MSE, RMSE, MAE) and fitting ability ($R^2$), providing a more accurate tool for deep learning resource scheduling.

## FUTURE WORK

Based on the progress made in GPU memory prediction in this study, future work will focus on deepening exploration in the following closely related directions: firstly, we will further optimize the model architecture to improve its efficiency and universality, explore more advanced attention mechanism variants or lightweight designs to reduce the complexity and inference latency of the model itself, making it more suitable for online resource scheduling scenarios with strict real-time requirements; At the same time, the plan is to conduct in-depth research on the generalization ability across tasks, frameworks, and hardware platforms. By constructing a large-scale heterogeneous dataset that includes a wider range of deep learning model architectures, different computing frameworks, and diverse GPU hardware, the system will evaluate and enhance the robustness and transferability of the model in complex and changing environments; In addition, we will focus on promoting the deep integration and application verification of the model in practical system ecosystems, such as embedding it into Kubernetes schedulers or cloud platform resource managers, developing intelligent memory aware scheduling strategies, achieving a closed loop from prediction results to resource decisions, and verifying its practical benefits in improving GPU utilization, reducing task queuing delays, and optimizing overall cost of ownership in real large-scale cluster production environments; Finally, considering the introduction of precise prediction of memory release mode, constructing a more complete life cycle portrait of memory, and exploring the combination of reinforcement learning technology, the model can not only passively predict demand, but also actively participate in and optimize training configuration strategies, ultimately forming an integrated intelligent memory management solution of "prediction decision optimization", providing core support for efficient and economical deployment of ultra large scale deep learning training and inference.